\documentclass[letterpaper, 10 pt, conference]{ieeeconf}  

\usepackage{amsmath,amsfonts}
\usepackage{algorithm}
\usepackage{array}
\usepackage[caption=false,font=normalsize,labelfont=sf,textfont=sf]{subfig}
\usepackage{textcomp}
\usepackage{stfloats}
\usepackage{url}
\usepackage{verbatim}
\usepackage{graphicx}
\usepackage{cite}
\usepackage{algpseudocode}
\usepackage{multirow}
\usepackage{hyperref}

\IEEEoverridecommandlockouts                              

\title{\LARGE \bf
An Efficient Projection-Based Next-best-view Planning Framework for Reconstruction of Unknown Objects
}

\author{Zhizhou Jia,  Shaohui Zhang and Qun Hao 
\thanks{This work has been submitted to the IEEE for possible publication. Copyright may be transferred without notice, after which this version may no longer be accessible.}
\thanks{All authors are with School of Optics and Photonics, Beijing Institute of Technology, Beijing 100081, China
        {\tt\small \{jiazhizhou, zhangshaohui, qhao\}@bit.edu.cn}}%
}

\begin{document}

\maketitle
\thispagestyle{empty}
\pagestyle{empty}

\begin{abstract}
Efficiently and completely capturing the three-dimensional data of an object is a fundamental problem in industrial and robotic applications. 
The task of next-best-view (NBV) planning is to infer the pose of the next viewpoint based on the current data, and gradually realize the complete three-dimensional reconstruction. 
Many existing algorithms, however, suffer a large computational burden due to the use of ray-casting. 
To address this, this paper proposes a projection-based NBV planning framework. 
It can select the next best view at an extremely fast speed while ensuring the complete scanning of the object. 
Specifically, this framework refits different types of voxel clusters into ellipsoids based on the voxel structure.
Then, the next best view is selected from the candidate views  using a projection-based viewpoint quality evaluation function in conjunction with a global partitioning strategy.
This process replaces the ray-casting in voxel structures, significantly improving the computational efficiency.
Comparative experiments with other algorithms in a simulation environment show that the framework proposed in this paper can achieve 10 times efficiency improvement on the basis of capturing roughly the same coverage. 
The real-world experimental results also prove the efficiency and feasibility of the framework.
\end{abstract}

\section{Introduction}
Efficient capturing the three-dimensional data of an object plays an important role in reverse engineering and quality inspection in industrial applications, as well as autonomous exploration and interaction in robotics applications.


In the actual scanning process, we will encounter the inability to completely reconstruct the object due to occlusion itself or the FOV limitation of acquisition device. 
In order to solve these problems, we need to combine the currently collected data to infer the pose of the camera's next viewpoint. This process is called "Next Best View" planning.

The NBV planning issue was first raised by Connoll in 1985\cite{1087372}. His method of using voxels to represent the object has been widely used and optimized by subsequent researchers. Currently, voxel structure is the most widely used representation structure in the field of NBV planning.
Directly calculating a six-degree-of-freedom pose from 3D space is quite challenging. Therefore, a common approach among researchers is to sample a large number of candidate viewpoints within a potential area and select the best viewpoint from them.
On the basis of voxel structures, researchers commonly employ ray-casting to assess the visibility of each voxel from the candidate viewpoints. This information is fed into a viewpoint quality evaluation function to select the most suitable viewpoint.

However, as the volume of the object increases and the demand for voxel precision rises, the ray-casting method faces a significant computational challenge.
The ray-casting method projects a series of rays from the optical center of the camera. 
During execution, the method determines the occlusion conditions in front of each voxel along the path of every ray. 
This process involves a large number of queries and calculations, thus the computational cost is considerably high.

\begin{figure}[t]
    \centering
    \includegraphics[width=3.4in]{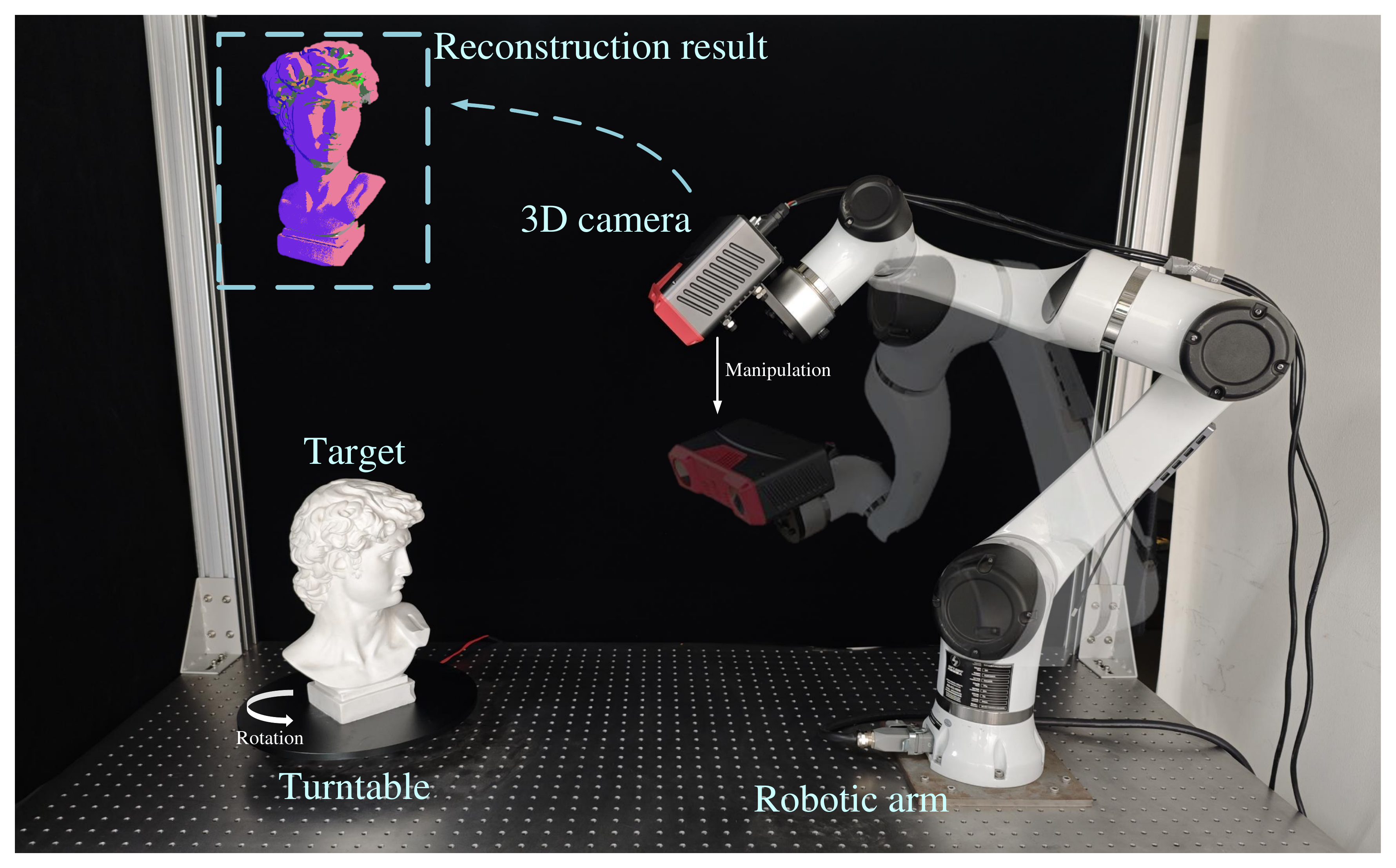 }
    \caption{Overview of the NBV planning experiment platform. An object to be measured is placed on a turntable, and a 3d camera is equipped at the end of the robotic arm for data acquisition. Our NBV planning framework accomplishes a complete reconstruction of the object by controlling the robotic arm and the turntable.\href{https://drive.google.com/file/d/1yQo_oHEjIK4X6LQc-lpWSFYunienocEl/view?usp=drive_link}{Video of the experiments is available.} }
    \label{Experimental platform}
\end{figure} 

To address this problem, this paper proposes a Projection-Based NBV Planning Framework based on the voxel structure. 
It can rapidly infer the pose of the next viewpoint while ensuring a complete reconstruction of the object.
We adopt Octomap\cite{hornung13auro} to classify all voxels as “occupied”, “frontier”, “unknown”, and "none". 
On this basis, the “occupied” and “frontier” voxels are clustered, and the clustered voxels are fitted into ellipsoids.
We integrate frontier information with ellipsoids and select the most suitable viewpoint through a projection-based viewpoint quality evaluation function.
To prevent backtracking caused by greedily selecting the locally optimal viewpoint, a global partitioning strategy has also been incorporated into the framework.
Comparing with the current SOTA algorithm based on voxel structure in simulation experiments, our framework can obtain 10 times efficiency improvement on the basis of roughly the same coverage. 
In addition, sufficient experiments have been done in real-world scenarios to prove the feasibility of the proposed framework. The real-world environment is shown in Fig. \ref{Experimental platform}.


In summary, the main contributions of this paper are the following three:

1) We propose a structure for representing
unknown object based on ellipsoid fitting.
  
2) We propose a projection-based viewpoint quality evaluation function to replace ray-casting, significantly enhancing the efficiency of framework.
  
3) We introduce a global partitioning strategy to avoid backtracking caused by greedy selection.

\section{Related Works}

The original goal of the NBV planning problem was to completely reconstruct a single object\cite{1087372}. 
As the problem has been studied in depth, researchers have expanded the concept of "object" to the entire environment. 
Therefore, in recent years, the NBV planning problem has become a hot research area, including both complete reconstruction of objects and environmental exploration. 
In this context, using voxel structures to represent objects is the most widely used method in current NBV planning problem.
Within the framework of voxel structures, current strategies for assessing viewpoint quality are broadly divided into three categories: based on frontier information, based on information entropy, and based on utility functions.

The viewpoint evaluation strategy based on frontier information was first proposed by Yamauchi\cite{613851} for mobile robots to explore unknown environments in 2D grid maps. 
Its core strategy is to use the frontier voxels around the candidate viewpoint to evaluate the information gain of the candidate viewpoint.
Since it can mark the optimal observation area in a global scope, it is widely used in the exploration of the entire scene\cite{9387089, 7487281, deng2020frontier, 9324988, 10423847}.

The viewpoint evaluation strategy based on information entropy was first proposed by Potthast\cite{POTTHAST2014148}.
Its core strategy is to employ ray-casting to identify the voxels along the same ray path.
Then, based on the occlusion conditions between voxels, different occupancy probabilities are assigned to each voxel.
Under this strategy, the potential information gain for each candidate viewpoint can be precisely determined. 
So, it is often applied in the precise reconstruction of individual objects\cite{7835670,9635628,7487527}.

The viewpoint evaluation strategy based on the utility function widely applied in real robots for object reconstruction and environmental exploration.
The main purpose of this strategy is to consider the constraints of the real environment in which the robot operates.
However,it is challenging to directly describe the information gain for each candidate viewpoint.
Therefore, it is often used in conjunction with other strategies. 
Utility functions can compensate for metrics that are overly greedy, helping to avoid situations where the robot cannot reach certain viewpoints\cite{7487527,10143748,6943158}.

The aforementioned strategies all require the use of ray-casting to assess the visibility and occlusion of voxels from a specific viewpoint.
Some scholars have also conducted research on how to improve the efficiency of ray-casting algorithms.  
Vasquez-Gomez\cite{6569201} proposed a hierarchical ray-tracing strategy. 
This approach first uses coarse ray-tracing to eliminate invalid positions and then conducts fine ray-tracing only for the valid locations.
Batinovicp\cite{9695290} uses the Shadowcasting algorithm to replace ray tracing. 
The Shadowcasting algorithm distinguishes the idle area from the occluded area, and then recursively processes the idle area and the occluded area respectively. 
Both methods can save the overhead of the ray tracing algorithm to a large extent. In addition, some researchers use deep learning methods to design NBV algorithms\cite{MENDOZA2020224,vasquez2021next,10342226}.
The most obvious advantage of this method is that it can directly avoid the use of any form of ray tracing algorithm and can give the best viewpoint of the next frame at a very fast speed. 
However, due to the lack of relevant data sets in the field of NBV planning, such algorithms are difficult to generalize to other unknown objects.


Inspired by previous research and the 3D Gaussian Splatting method\cite{10.1145/3592433}, this paper proposes an NBV planning framework based on voxel structures. 
This framework assesses the potential observation quality of candidate viewpoints by fitting frontier voxels into an ellipsoid and evaluating them through projection.
It completely replaces ray-casting, significantly improving the computational efficiency. 
The next section will elaborate on the specifics of this framework in detail.

\section{Methodology}
\subsection{Framework Overview}

\begin{figure*}[!ht]
    \centering
    \includegraphics[width=7in]{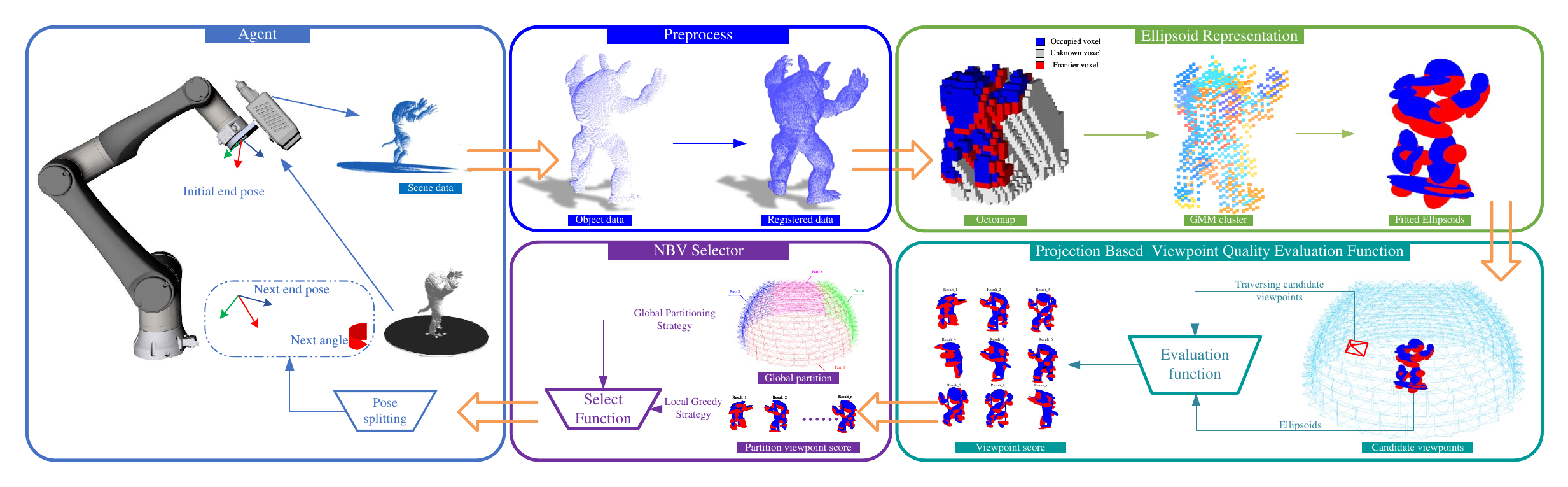}
    \caption{Overview of NBV Planning framework proposed in this paper. The orange arrows describe the running steps of the NBV iteration process.}
    \label{System Overview}
\end{figure*}

The projection-based NBV planning framework proposed in this paper will first represent the collected data on the voxel structure.
It categorizes voxels into different types and then fits them into ellipsoids. Subsequently, it employs a projection-based viewpoint quality evaluation function integrated with a global partitioning strategy to select the next viewpoint for the next frame among the candidate viewpoints. The entire workflow is depicted in Fig \ref{System Overview}.
First, the robot arm will move to an artificially determined initial observation pose.
When the robot arm reaches this pose, the 3D camera will collect a frame of point cloud data. 
After preprocessing, the data is passed to the NBV planning framework, thereby starting the iterative process of completely reconstructing the unknown object.
\subsection{Proposal of Candidate Viewpoints}
In conjunction with the experimental platform shown in Fig. \ref{Experimental platform}, this paper assumes that the radius of the bounding box of the object to be reconstructed will not exceed the distance from the robotic arm base to the turntable. 
Considering the depth of field limitation of the 3D camera, we propose a strategy to dynamically adjust the candidate view sampling area to ensure that the object can be completely reconstructed while meeting the optimal shooting distance of the 3D camera.

\begin{figure}[!h]
    \centering
    \includegraphics[width=3.4in]{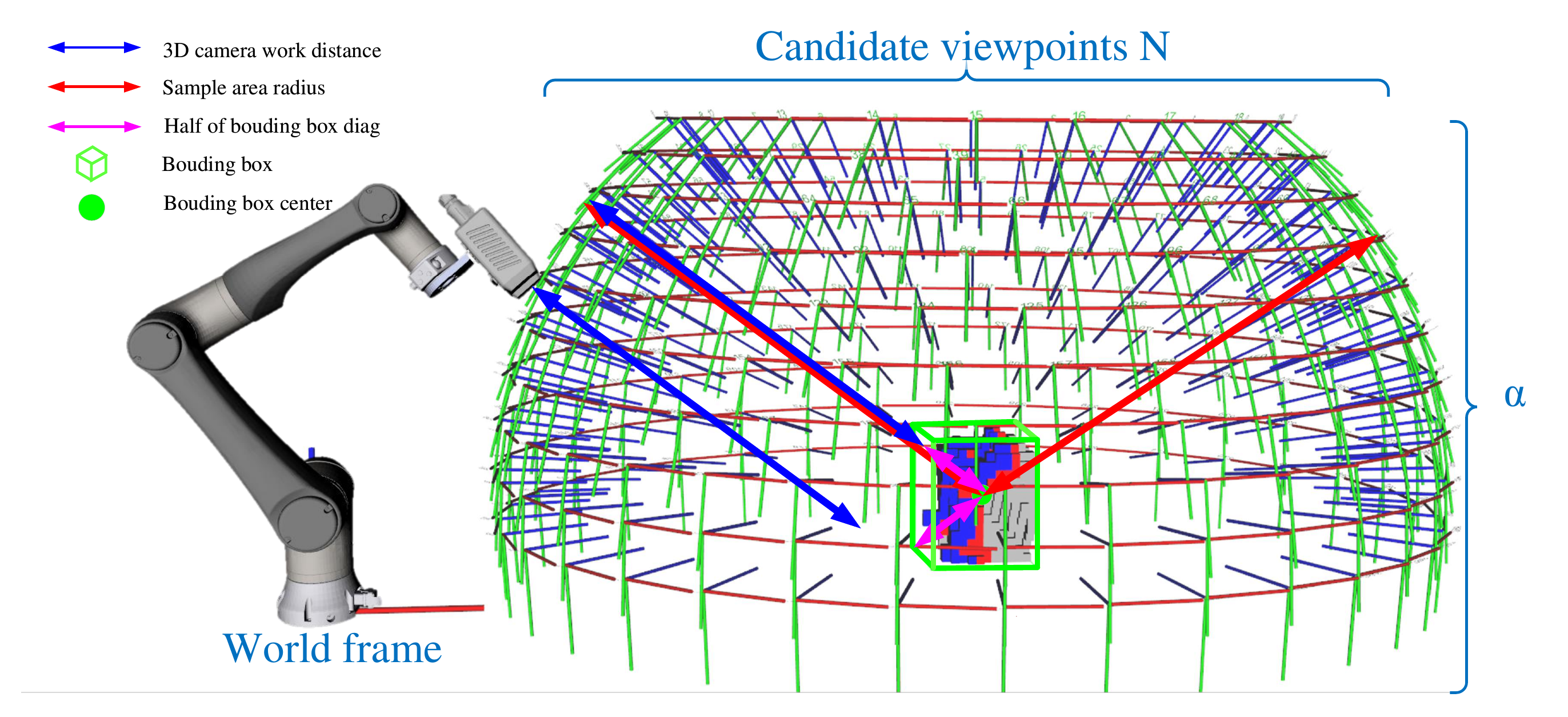}
    \caption{The components of the radius of the candidate viewpoint sampling region and the results of candidate viewpoint sampling within this region.}
    \label{Proposal of Candidate Viewpoints}
\end{figure}

Since the reachability of the robot arm must be considered, we set the sampling area of the candidate viewpoint to the partial hemisphere area above the turntable.
The algorithm uses the base coordinates of the robot as the world coordinate system, as shown in Fig. \ref{Proposal of Candidate Viewpoints}. 
The radius of the hemisphere is $R = d_c + d_b$, where $d_c$ is the working distance of the 3D camera and $d_b$ is half the diagonal length of the unknown object bounding box. 
During the reconstruction process, the bounding box of the unknown object will change, so the size of the hemisphere will also be dynamically adjusted to adapt to the depth of field limit of the 3D camera.
From the hemispherical candidate area, we first selected $\alpha$ equidistant parallels.
Then, according to the length of each parallel, $N$ candidate viewpoints are distributed proportionally along each parallel. 
All viewpoints are oriented uniformly towards the center of the bounding box.

\subsection{Voxel Structure construction}
\begin{figure}[!ht]
    \centering
    \subfloat[]{\includegraphics[width=1.6in]{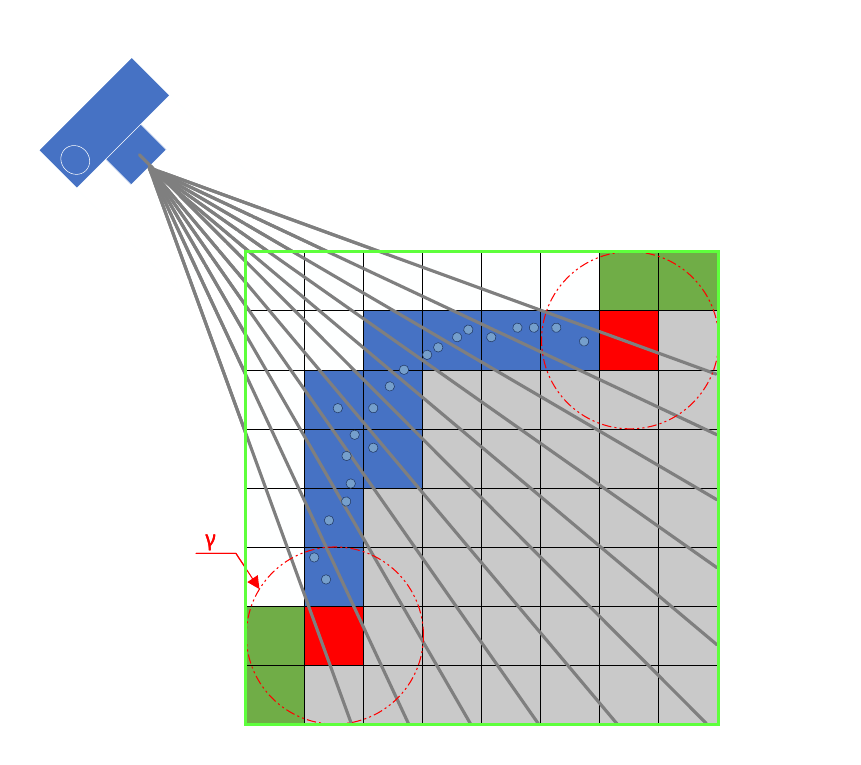}
    \label{Voxel struct(a)}
    }
    \subfloat[]{\includegraphics[width=1.6in]{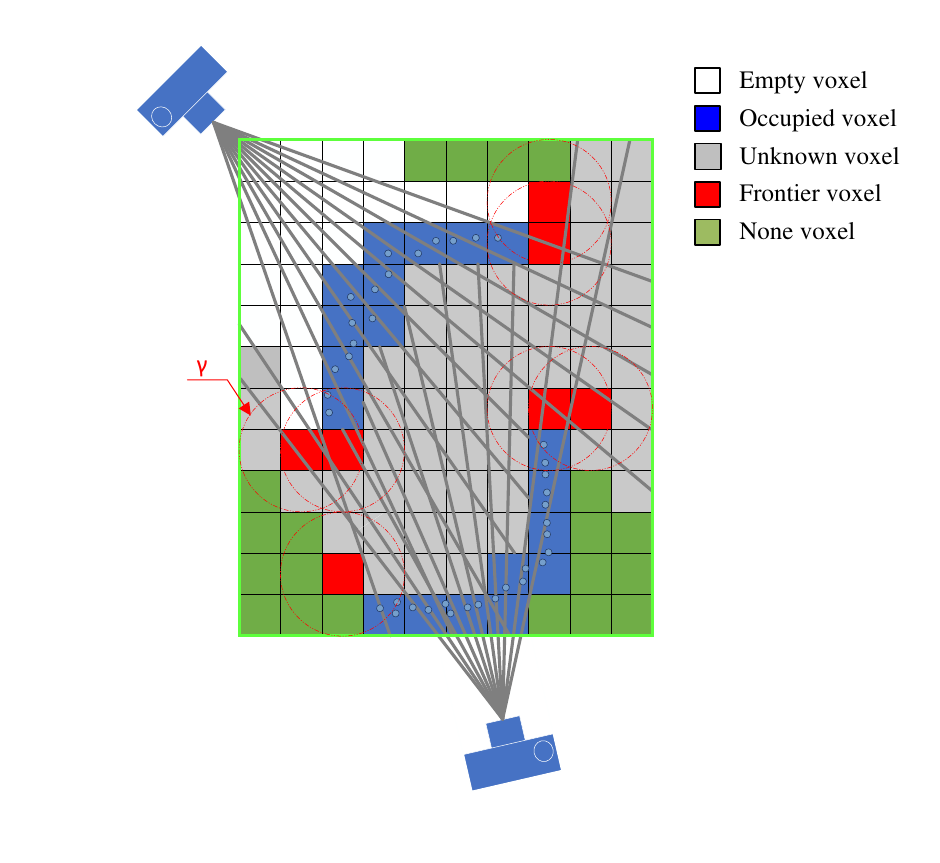}
    \label{Voxel struct(b)}
    }
    \caption{Use a 2D grid to describe the classification rules of voxels in Octomap. (a) Classification results of one frame input. (b) Classification results of multiple frames input. The green box is the bounding box of the object.}
    \label{Voxel struct}
\end{figure}

Once the 3D camera captures a frame of point cloud data, it is first subjected to preprocessing.
Subsequently, the preprocessed point cloud is registered into the point cloud set $P_f$.
Finally, the most recently registered point cloud $P_f$ is subjected to voxel construction. 
This paper divides the voxels in the unknown object bounding box into five categories, namely Empty $V_e$, Occupied $V_o$, Unknown $V_u$ Frontier $V_f$ and None $V_n$, corresponding to the white, blue, gray, red and green squares in Fig. \ref{Voxel struct}.

Each time the bounding box is updated, the newly added voxels are initially set to $V_n$.
$V_o$ represents the surface structure of the unknown object, which is the voxel occupied by $P_f$.
$V_e$ represents the free area in the bounding box, which is the set of voxels not occupied by $P_f$ after being scanned by the acquisition device.
$V_u$ represents the area where the surface structure of the unknown object may exist. 
These areas are not captured by the acquisition device because they are blocked by $V_o$. 
As shown in Fig. \ref{Voxel struct}\subref{Voxel struct(a)}, With each new addition of occupied voxels $V_o^i$, $V_u^i$ and $V_e^i$ will be generated accordingly.

This paper uses the ray-trave method proposed by Amanatides\cite{amanatides1987fast}. 
After sampling several rays at equal angles in the FOV area of the current viewpoint, we define all $vi \in V_o$ that the ray passes through before $V_o^i$ as $V_e^i$, and all $vi \in V_o$ that the ray passes through after $V_o^i$ as $V_u^i$. 
Once $V_e$ and $V_o$ are determined, they will not change in subsequent updates.

The update of $V_f$ is performed after the update of $V_u$ and $V_o$. 
Any $v_{i}\in V_u$ is considered a Frontier voxel as long as there are both empty and occupied voxels in its spatial neighborhood, 
that is:
\begin{equation}
    V_f = \{v_u^{i} \mid \exists  v_e \in \mathcal{N}(u^{i}) \And v_o \in \mathcal{N}(u^{i})  \}
\end{equation}
where $v_u^{i} \in V_u$ and $\mathcal{N}(u^{i}) $represents the spatial neighborhood of $v_u^{i}$.


The bounding box $\mathbf{B}$ is initialized to the bounding box of $V_o$ in the first frame.
Then,$\mathbf{B}$ is expanded along the direction of the current viewpoint, doubling the diagonal of to twice its current length.
In subsequent updates, a sphere $s_i$ with a radius of $\gamma$ is generated around each voxel $v_i \in V_f$, and the size of $\mathbf{B}$ is expanded to cover $V_o$ and $s_i$. 
The result is shown in Fig. \ref{Voxel struct}\subref{Voxel struct(b)}.

\subsection{Ellipsoid Representation}

\begin{figure}[h]
    \centering
    \includegraphics[width=3.4in]{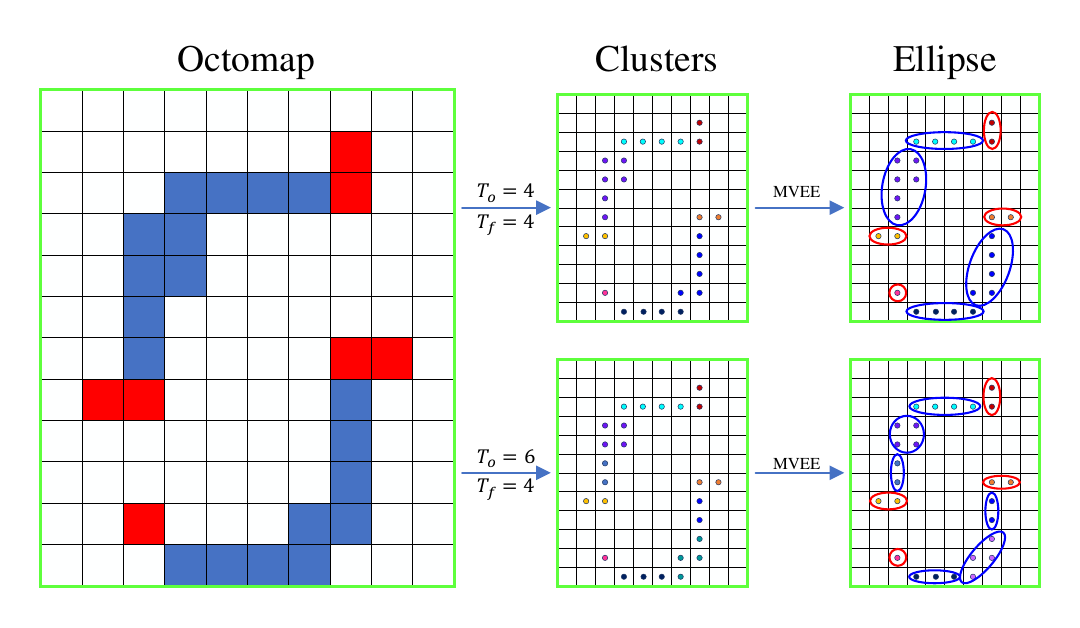}
    \caption{Results of different GMM clustering numbers. \( T_o \) represents the number of clusters of Occupied voxels, and \( T_f \) represents the number of clusters of Frontier voxels.}
    \label{Ellipsoid representation}
\end{figure}

In 3D games, the collision structure of an entity can be represented by an ellipsoid\cite{fauerby2003improved}. 
If a more sophisticated collision structure is required, multiple ellipsoids can be used to represent different parts of an entity.
Inspired by this, this paper also uses multiple ellipsoids to describe a object structure. 
Since the object in this paper is unknown, its structural characteristics cannot be clearly determined. 
Therefore, it is necessary to roughly divide the object into \( t \) clusters, and fit each cluster into an ellipsoid to represent the object structure. 
Fig. \ref{Ellipsoid representation} illustrates this process using a two-dimensional diagram

Inspired by 3DGS\cite{10.1145/3592433}, the equiprobability density surface of 3D Gaussian distribution is an ellipsoid. 
If a group of scattered points in space conform to the 3D Gaussian distribution, its covariance matrix can reflect the degree of dispersion of these points in each axis direction, and it is easy to fit this cluster of points into an ellipsoid. 
Therefore, this paper will use the Gaussian mixture model (GMM) to cluster voxels.

Occupied voxels and frontier voxels will directly affect the viewpoint quality evaluation results.
Therefore, after each update of $V_o$ and $V_f$, we will initialize $T$ 3D Gaussian distributions for this two types of voxels.
Then we use the expectation maximization (EM) algorithm to train the Gaussian distribution to obtain the results $C_o$ and $C_f$ after clustering these two types of voxels.

The selection of the number of Gaussian distributions $T$ will directly affect the number of fitted ellipsoids in the subsequent algorithm. 
Too few ellipsoids will make it difficult to accurately reflect the local structure of the object, and too many ellipsoids will burden the overall algorithm's operating efficiency. 
In order to enable the algorithm to adaptively select the number of Gaussian distributions, this paper introduces the Bayesian Information Criterion (BIC)\cite{c4048c8f-6ca9-3965-96a3-653ab8996955} to evaluate the model parameters.

\begin{equation}
    \text{BIC} = k \ln (n) - 2 \ln *(L)
\end{equation}

$k$ is the number of model parameters, $n$ is the number of samples, and $L$ is the likelihood function.

The core idea of BIC is to find the optimal balance between the number of parameters in the model and the value of the likelihood function. 
When an increase in model parameters leads to a significant improvement in the likelihood function value, a larger model is chosen; otherwise, a smaller model is selected. 
In order to balance efficiency and accuracy, this paper sets the value range of T to [5, 50] during the iteration process. Among the two voxels $V_o$, $V_f$, select $T_o$, $T_f$ with the smallest corresponding BIC value as the number of Gaussian distributions in the current iteration, that is:
\begin{equation}
    T_{*} = \arg \min_{T_{max}}BIC(GMM(V,T))
\end{equation}


After obtaining $C_o$ and $C_f$, we use the method proposed by Gärtner\cite{gartner1997smallest} in the CGAL library to calculate the minimum volume closed ellipsoid model (MVEE).
The principle of its implementation is that for each point $q$ in a single cluster $c_i \in C_{*}$, recursively calculate the minimum enclosing ellipsoid $E$ of the point set $c^i \setminus \{q\}$ after removing the point. 
If the point $q$ is within the currently calculated ellipsoid $E$, the current ellipse is the minimum enclosing ellipse. 
If the point $q$ is not within the ellipse $E$, the ellipse needs to be updated to include the point $q$. 
This step is expressed as:
\begin{equation}
    E_*^i = \text{SmEll}(C_*^i \setminus \{q\}, \{q\})
\end{equation}

Through the aforementioned process, we can perform ellipsoid fitting on $C_o$ and $C_f$. Consequently, we obtain $E_o$ and $E_f$, representing the fitting results of the Occupied voxels and Frontier voxels.
At this point, we can transform the representation of the target from a voxel structure to ellipsoid structure.
Each ellipsoid is described by its own equation.

\subsection{Projection-based Viewpoint Quality Evaluation Function}

The goal of the viewpoint quality evaluation function proposed in this paper is to enable the viewpoint to observe the most frontier information. 
In this process, we must consider the occlusion problem between structures when observing the object.
Although ray-casting can accurately give the occlusion relationship of each voxel, the extensive use of ray-casting will take up a lot of computing resources.
To address this, we use a new strategy that calculates the observable weight by using the order of the center positions of each ellipsoid under the current viewpoint. 
Then obtains the final viewpoint evaluation result by accumulating the results of separate weighted projections of each ellipsoid.
The details of evaluating a single viewpoint can be divided into three parts: ellipsoid observation weight calculation, ellipsoid weighted projection calculation, and viewpoint quality result calculation.

\begin{figure*}[!t]
    \centering
    \subfloat[]{\includegraphics[width=1.1in]{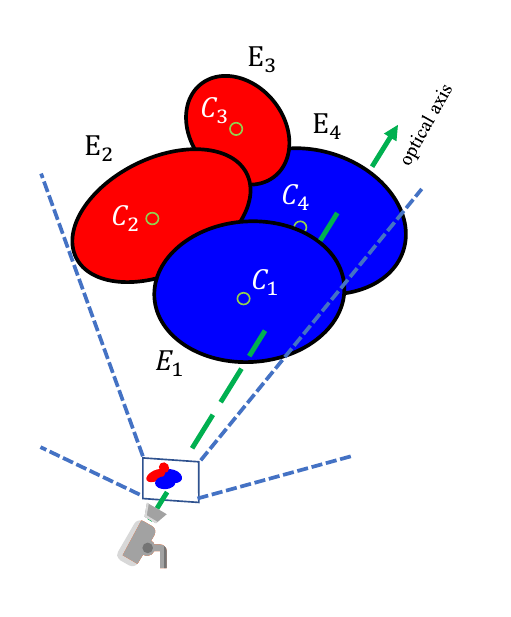}
    \label{Viewpoint quality evaluation function(a)}
    }
    \subfloat[]{\includegraphics[width=2.2in]{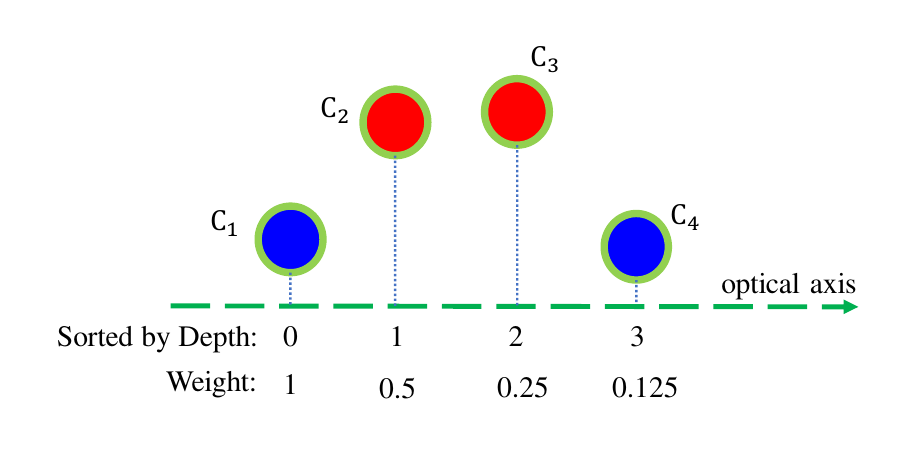}
    \label{Viewpoint quality evaluation function(b)}
    }
    \subfloat[]{\includegraphics[width=2.2in]{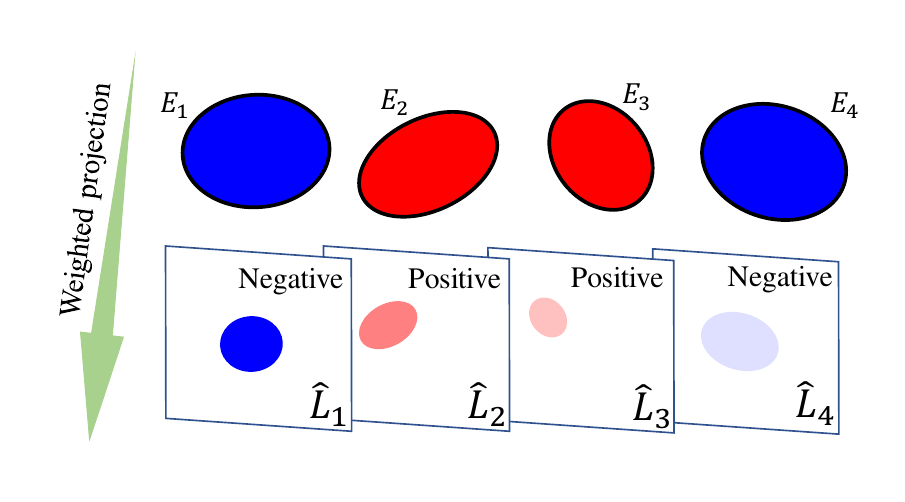}
    \label{Viewpoint quality evaluation function(c)}
    }
    \subfloat[]{\includegraphics[width=1.5in]{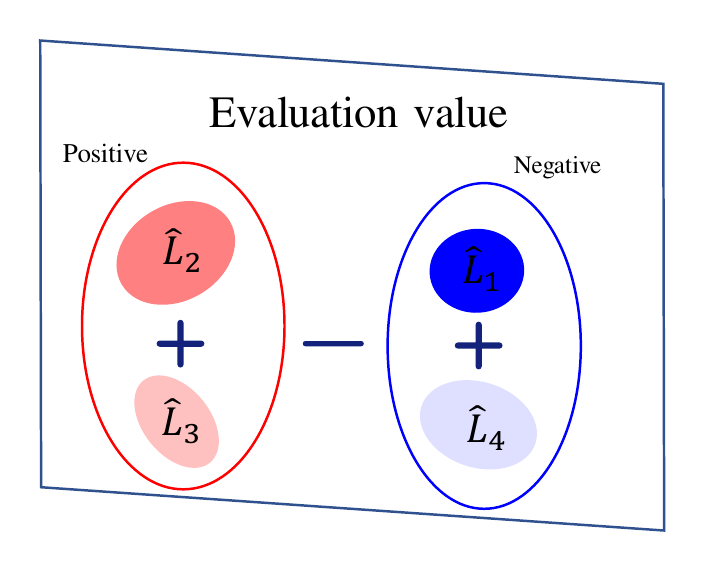}
    \label{Viewpoint quality evaluation function(d)}
    }
    \caption{Details of the quality evaluation for a single viewpoint. (a) Convert the center point of each ellipsoid to the current viewpoint coordinate system. (b) Sort the ellipsoids according to the depth of the center position and assign observability weights. (c) Project each ellipsoid based on the observability weight to obtain the observability measure. (d) Combine the Occupied ellipsoid and the Frontier ellipsoid to give the current viewpoint quality evaluation value.}
    \label{Viewpoint quality evaluation function}
\end{figure*}

\subsubsection{Ellipsoid observation weight calculation}

As shown in Fig. \ref{Viewpoint quality evaluation function}\subref{Viewpoint quality evaluation function(a)}.
In the obtained ellipsoid $E$, we can obtain the center coordinates $C_w$ of the ellipsoid in the world coordinate system. 
On this basis, according to the rotation matrix $R$ and translation matrix $t$ of the current candidate viewpoint relative to the world coordinate system, we can calculate the center $C$ of the ellipsoid in the coordinate system of the candidate viewpoint.
\begin{equation}
    C = R^{-1}(C_w-t)
\end{equation}
As shown in Fig. \ref{Viewpoint quality evaluation function}\subref{Viewpoint quality evaluation function(b)}, We sort all ellipsoid centers $C$ in ascending order according to their z-coordinates.
Each ellipsoid is assigned a unique depth rank $r$.
After obtaining the ranks, we calculate the observable weight $W$ for each ellipsoid as follows:
\begin{equation}
    W = 0.5^{r}
\end{equation}
\subsubsection{Ellipsoid weighted projection calculation}
Any ellipsoid $E$ in space can be described by a matrix $Q$, that is:
\begin{equation}
    X^TQX=0
\end{equation}
Where $X=[x \ y \ z \ 1]^{T}$.
For any given ellipsoid, the ellipse it projects onto under the action of the camera projection matrix $P$ can be represented by the matrix description $\Phi$.
\begin{equation}
    \Phi^* = PQ^*P^T
\end{equation}
Where $Q^*$ is the dual matrix of $Q$, $\Phi^*$ is the dual matrix of $\Phi$. 
If any matrix $M$ is a reversible matrix, its dual matrix is $M^*=M^{-1}$. 
The camera matrix $P$ can be calculated by the internal parameter $K$ of the current 3D camera, and the rotation matrix $R$ and translation matrix $t$ of the current candidate viewpoint relative to the world coordinate system.
\begin{equation}
    P = K[R^{-1}|R^{-1}t]
\end{equation}
As shown in Fig. \ref{Viewpoint quality evaluation function}\subref{Viewpoint quality evaluation function(c)}.
After acquiring the projected ellipse equation of ellipsoid $E$ from the current viewpoint, we proceed to project each ellipsoid individually onto the 3d camera's imaging plane.
Then, we calculate the sum of pixel values $L$ within each elliptical region in the image and multiply it by the observability weight $w$ to obtain a measure of the ellipsoid's observability, denoted as $\hat{L}$.

\begin{equation}
    \hat{L} = Lw 
\end{equation}
\subsubsection{Viewpoint quality evaluation function}
In order to allow the viewpoint to observe the maximum amount of frontier information, it is necessary to prioritize the projection of larger frontier ellipsoids over occupied ellipsoids.
For this purpose, we designed a simple evaluation function to achieve this goal:
\begin{equation}
    F = \sum_{E_f} \hat{L}(E_i) - \sum_{E_o}\hat{L}(E_i)
\end{equation}
As shown in Fig. \ref{Viewpoint quality evaluation function}\subref{Viewpoint quality evaluation function(d)}.
The projection of the frontier ellipsoid is the positive part of the observation quality assessment, while the projection of the occupied ellipsoid is the negative part.
Therefore the sum of the weighted projections of all frontier ellipsoids minus the sum of the weighted projections of the occupied ellipsoids can represent the measurement of the size of the frontier projection in the front of the viewpoint.
This approach satisfies our requirements for viewpoint quality evaluation.

\subsection{Global partitioning strategy}
In the framework proposed in this paper, ICP\cite{121791} is needed to perform frame-by-frame registration of the point cloud to correct the pose. 
If the viewpoint with the highest observation quality $F$is selected greedily as the best viewpoint, it can lead to discontinuous scanning of the target, which may result in the failure of point cloud registration.
Moreover, an overly greedy selection strategy can also lead to backtracking problem.
\begin{figure}[h]
    \centering
    \includegraphics[width=3.4in]{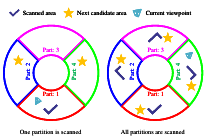}
    \caption{Results of different GMM clustering numbers. \( T_o \) represents the number of clusters of Occupied voxels, and \( T_f \) represents the number of clusters of Frontier voxels.}
    \label{Global partitioning strategy}
\end{figure}

To address this, this paper introduces a global partitioning strategy.
We divides the candidate view sampling area into $\beta$ regions according to longitude.
The selection of $\beta $ depends on the current registration strategy. 
After all regions have been scanned, the next best view can be selected from all partitions with the largest observation quality $F$.
Otherwise, next best view can only be selected in the unscanned neighborhood of the scanned partition, as shown in Fig. \ref{Global partitioning strategy}.

So far, this paper describes in detail the calculation process of the next best viewpoint after a frame of point cloud data is input. 
The process of completely reconstructing an object is shown in Algorithm \ref{algorithm}.
\begin{algorithm}
    \caption{Projection-Based NBV Planning Framework}
    \begin{algorithmic}[1]
        \State \textbf{Input:}agent pose $\xi$, $\alpha$, $N$, $\gamma$, $d_c$, $T_{max}$, $T_{min}$, $\beta$, $K$
        \State \textbf{Output:}target point cloud $\mathbf{\psi}$
        
        \While{Not Terminated}
            \State $\text{AgentExcute}(\xi)$
            \State $P \gets \text{TrigCamera}()$
            \State ${\mathbf{\psi}} \gets \text{Preprocess}(P)$
            \State $V_o, V_f, \mathbf{B} \gets \text{VoxelStructConstruct}(\mathbf{\psi}, \gamma)$
            \State $E_o, E_f \gets \text{EllipsoidsFit}(V_o, V_f, T_{min},T_{max})$
            \State $\mathbf{S} \gets \text{ProposeCandidateViewpoints}(\alpha, N, d_c, \mathbf{B})$
            \For{$s_i \in S$}
                \If {$ \text{InValidPartion}(s_i, \beta)$}
                    \State $F_i \gets \text{QualityEvaluate}(E_o, E_f, s_i, K)$
                \Else
                    \State $F_i \gets -\inf$
                \EndIf
            \EndFor
            \State $\xi \gets \text{SelectNBV}(\mathbf{S},\mathbf{F})$
        \EndWhile
    \end{algorithmic}
    \label{algorithm}
\end{algorithm}

\section{Experiments}
In order to verify the efficiency and practicality of the framework proposed in this paper, sufficient experiments were conducted in both simulation and real-world environments.
\subsection{Simulation experiments}
\begin{figure}[h]
    \centering
    \subfloat[]{\includegraphics[width=1.5in]{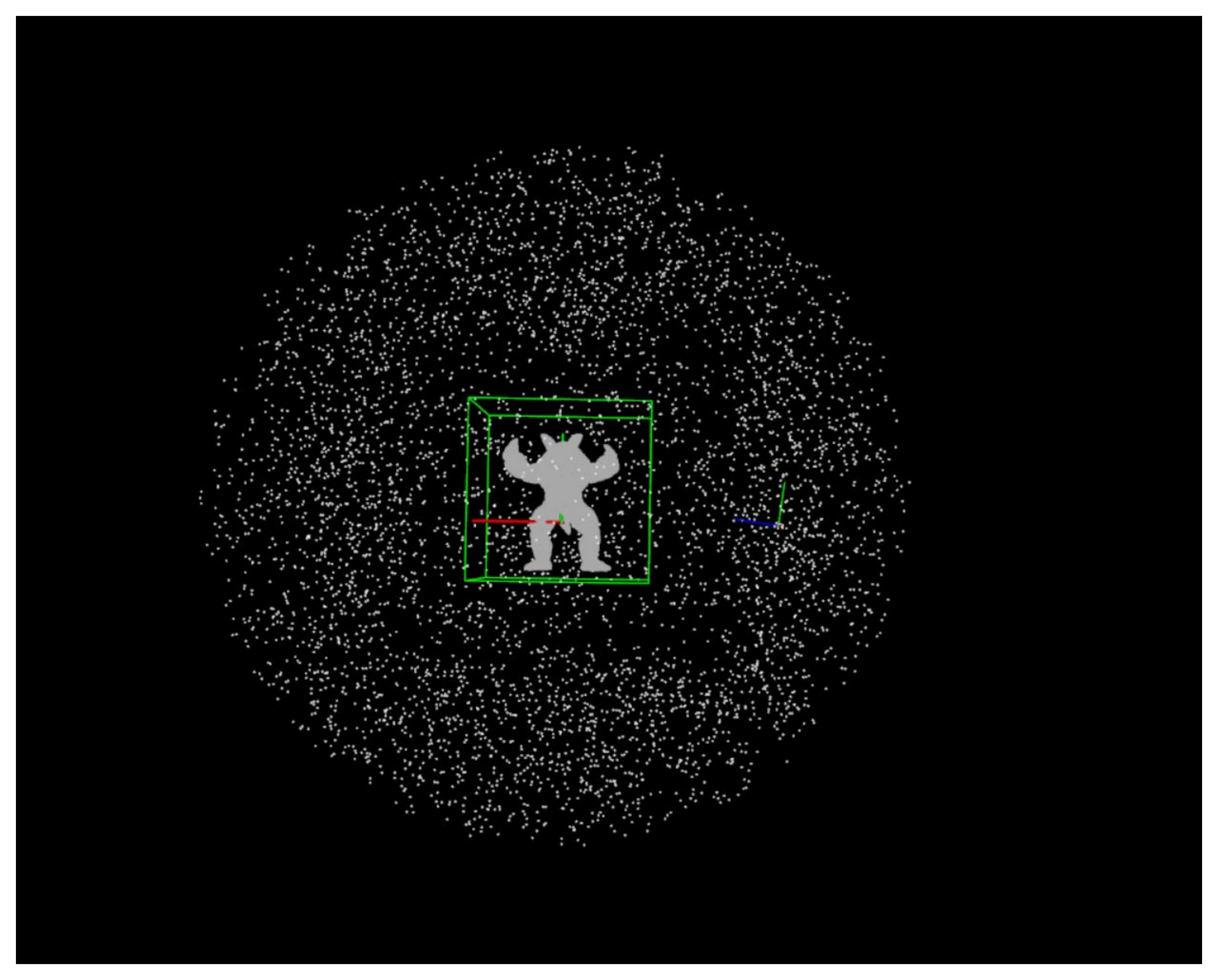}
    \label{Simulation environment(a)}
    }%
    \hfil
    \subfloat[]{\includegraphics[width=1.5in]{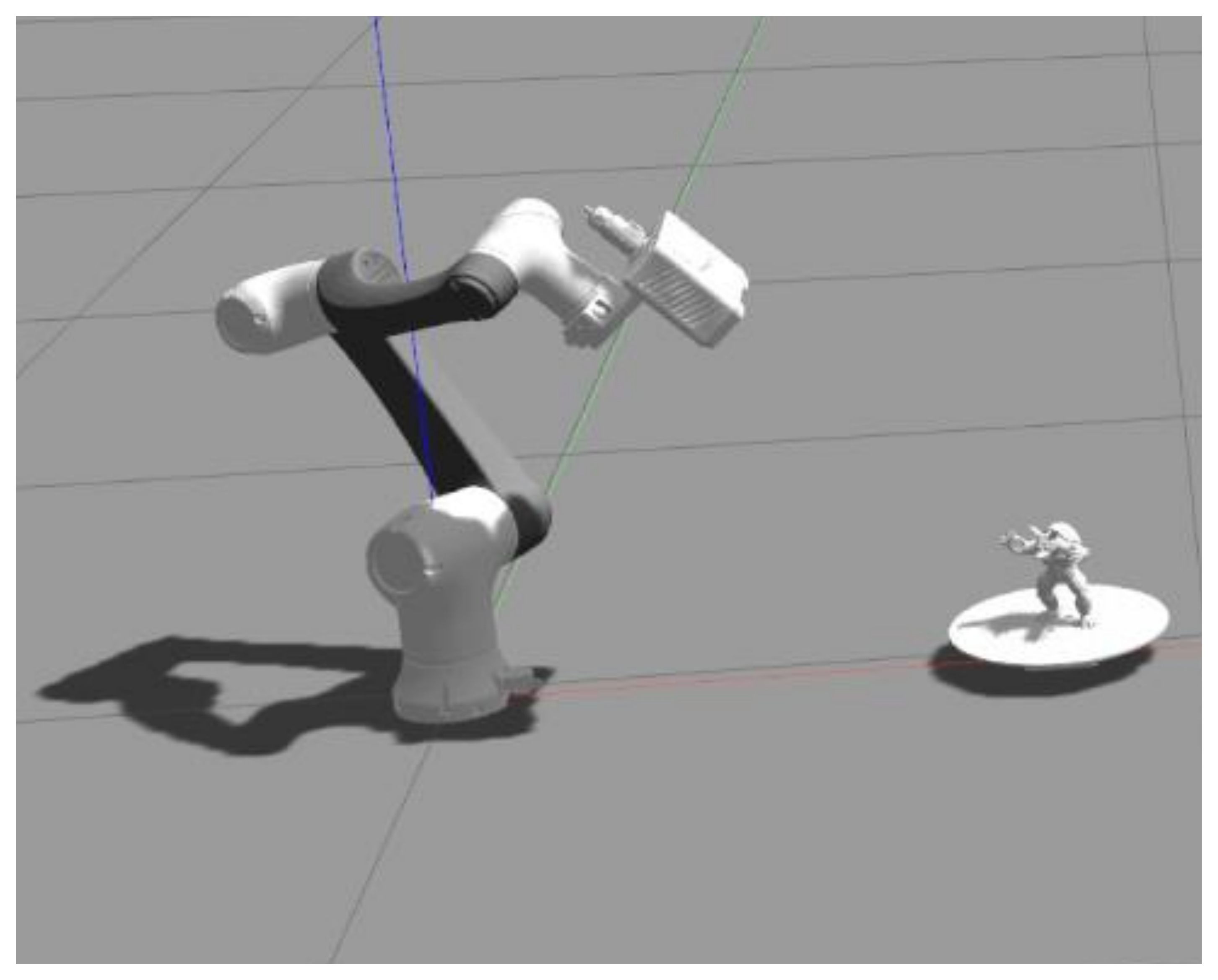}
    \label{Simulation environment(b)}
    }%
    \caption{Simulation environments for the comparative experiments. (a) The simulation environment for APORA and MFMR. (b) The simulation experiment environment used in this paper.}
    \label{Simulation environment}
\end{figure}

\begin{figure}[h]
    \centering
    \includegraphics[width=3.4in]{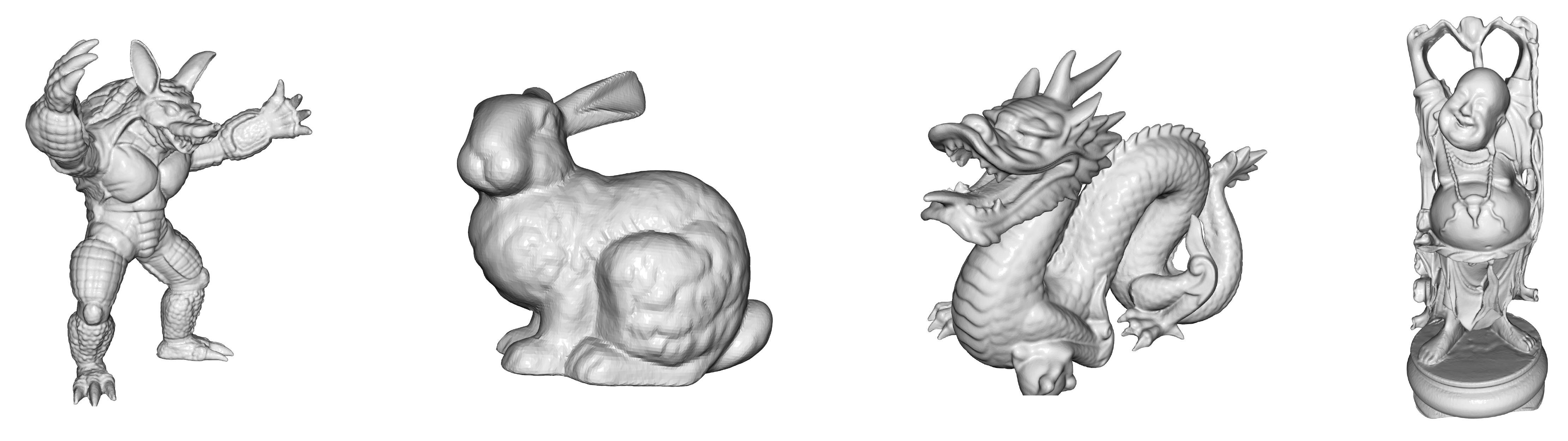 }
    \caption{Models used in the simulation experiments, from left to right: Armadillo, Bunny, Dragon, Buddha.}
    \label{Simulation target}
\end{figure}

In the simulation experiment, the algorithm proposed in this paper is compared with the SOTA algorithms APORA\cite{7835670} and MFMR\cite{9635628}. 
Both algorithms are deployed in the code provided by MFMR. 
The framework proposed in this paper is combined with the actual measurement scene, so the simulation scene used in this paper is different from the comparison algorithm. It uses Gazebo to create a close-to-real scene, as shown in Fig. \ref{Simulation environment}. 
In order to ensure the fairness of the comparison experiment, the input of all algorithms is the same size model from the Stanford 3D Scanning Repository, as shown in Fig. \ref{Simulation target}, and all experiments are performed in a PC with a processor of i7-14700f.

In the simulation experiments, the compared algorithms all used the same Octomap resolution of 0.01m. 
The number of partitions in this paper is set to 4,the number of candidate viewpoints is 400.
Although the intermediate processes of the planning algorithms differ, each algorithm received the same input data with ground truth. 
Therefore, two comparison metrics were used:
\begin{enumerate}
    \item Point Cloud Coverage: Sampling 10,000 points from the model file, traversing all model points, and calculating the proportion of model points that have a point in the input point cloud within a distance of 0.005m.
    \item Computational Efficiency: Measuring the time taken by the NBV planner to provide the next best view after receiving a frame input.
\end{enumerate}

\begin{figure}[h]
    \centering
    \includegraphics[width=3.4in]{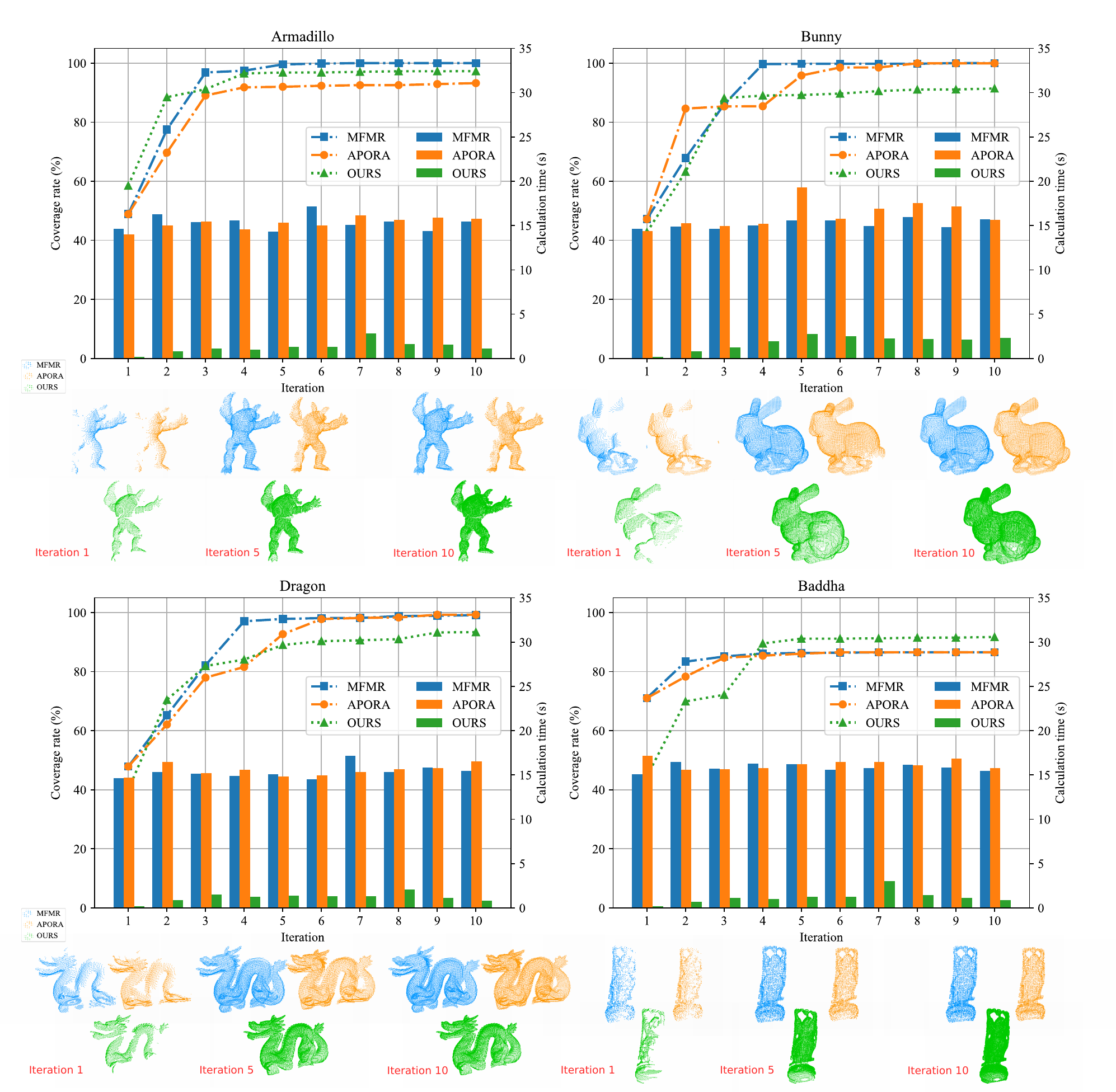 }
    \caption{Comparative results of the simulation experiments.}
    \label{Simulation result chart}
\end{figure}

\begin{table}[h]
    \caption{Total calculation running time (s)}
    \centering
    \begin{tabular}{llllll}
    \hline
      & Armadillo & Bunny & Dragon & Buddha & Average \\ 
    \hline
    MFMR      &153.729&151.721&153.394&158.577&15.436 \\ 
    APORA     &152.826&162.144&155.000&162.056&15.891\\ 
    OURS      &\textbf{12.605}&\textbf{18.254}&\textbf{11.767}&\textbf{11.858}&\textbf{1.320}\\ 
    \hline
    \end{tabular}
    \label{algorithm run time}
\end{table}

Fig. \ref{Simulation result chart} presents the simulation results. 
The line chart and bar chart respectively represent the point cloud coverage and calculation time of each iteration.
The scanning results as the iteration progresses are shown below the chart. 
The results indicate that the proposed method significantly improves computational efficiency while maintaining a similar coverage rate. 
It should be noted that in our simulation environment, the bottom of the object cannot be observed due to occlusion, while in the MFMR simulation environment the object can be observed from all viewpoints.
so the coverage of the model with a larger bottom in our results is slightly lower than that of MFMR is reasonable.

Table \ref{algorithm run time} provides the computational time data for different models and algorithms.
The last column shows the average time per iteration for different algorithms.
By calculating the average time consumed per iteration, the performance improvement of the proposed algorithm compared to other algorithms is quantified. 
Our first calculation is based on the manual specification of the structure and does not participate in the calculation of the average iteration time. 
Compared to MFMR and APORA, the proposed algorithm achieves performance improvements of \textbf{1069\%} and \textbf{1097\%}.

\subsection{Real-world experiments}

\begin{figure}[h]
    \centering
    \includegraphics[width=3.4in]{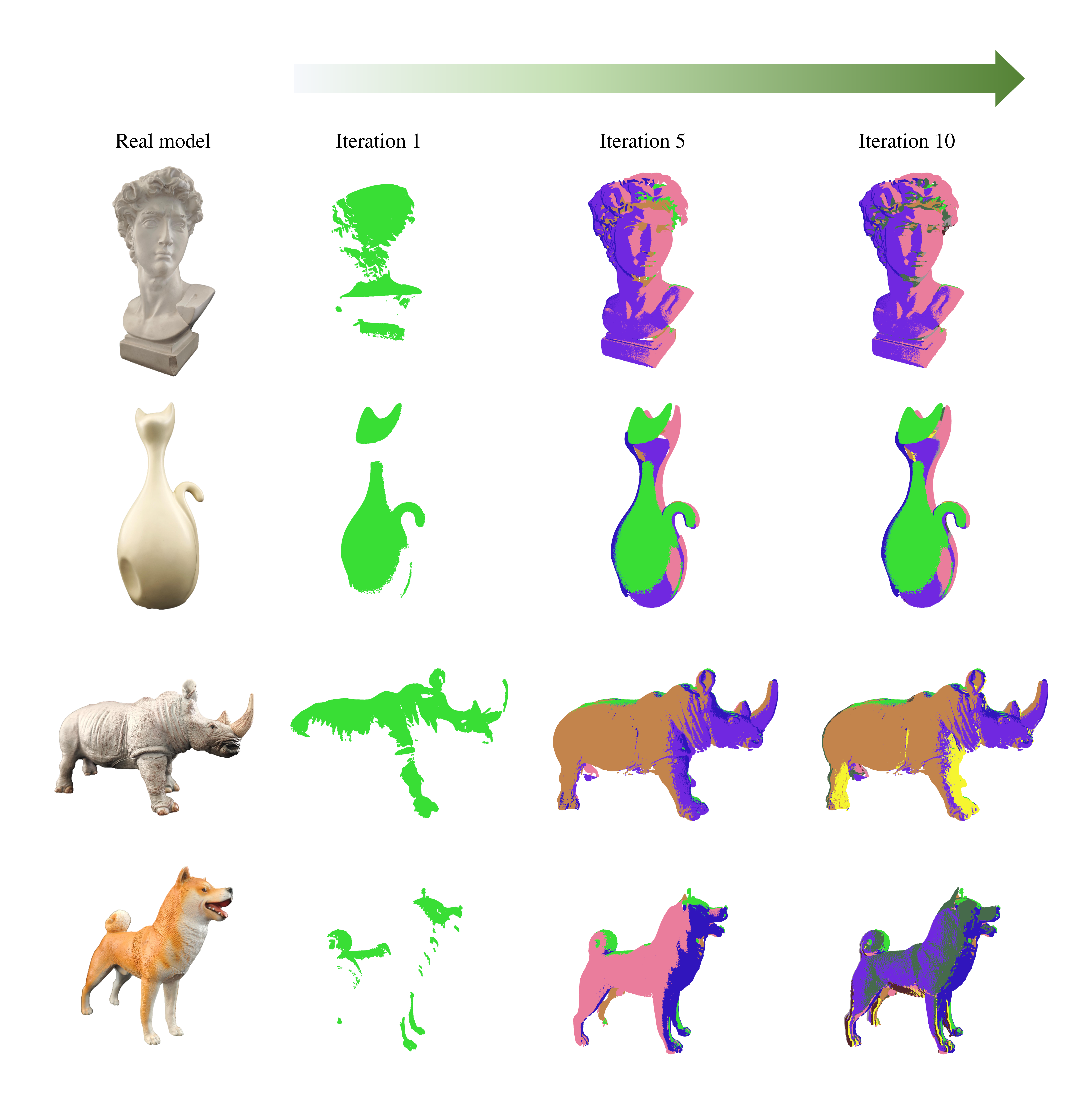}
    \caption{The models used in the real-world experiment and their reconstruction results, from top to bottom:  statue, vase, rhino and dog.}
    \label{Real-world result}
\end{figure}

The real-world experimental platform is depicted in Fig. \ref{Experimental platform}.
The data acquisition device is a self-made structured light 3d camera, and the robotic arm is the elfin5 from Hans Robotics, with a working range of 800mm. 
A 360-degree rotatable turntable is used to expand the relative position between the measured object and the robotic arm to the entire hemispherical space. 
To better perform data registration, we attached some highly reflective markers to the turntable. 
The NBV planning framework and various module control programs are all deployed on a laptop with an i9-14700HX processor. 
The real-world experiment also employed an Octomap resolution of 0.01m, with the number of partitions set to 4, and ten acquisitions will be made for each object to observe the reconstruction effect. 
This paper uses four targets of different sizes for reconstruction, as shown in Fig. \ref{Real-world result}, and the reconstruction results also demonstrate that the framework proposed in this paper can achieve good results when reconstructing different objects, proving the feasibility of the framework.

\begin{figure}[h]
    \centering
    \includegraphics[width=3.4in]{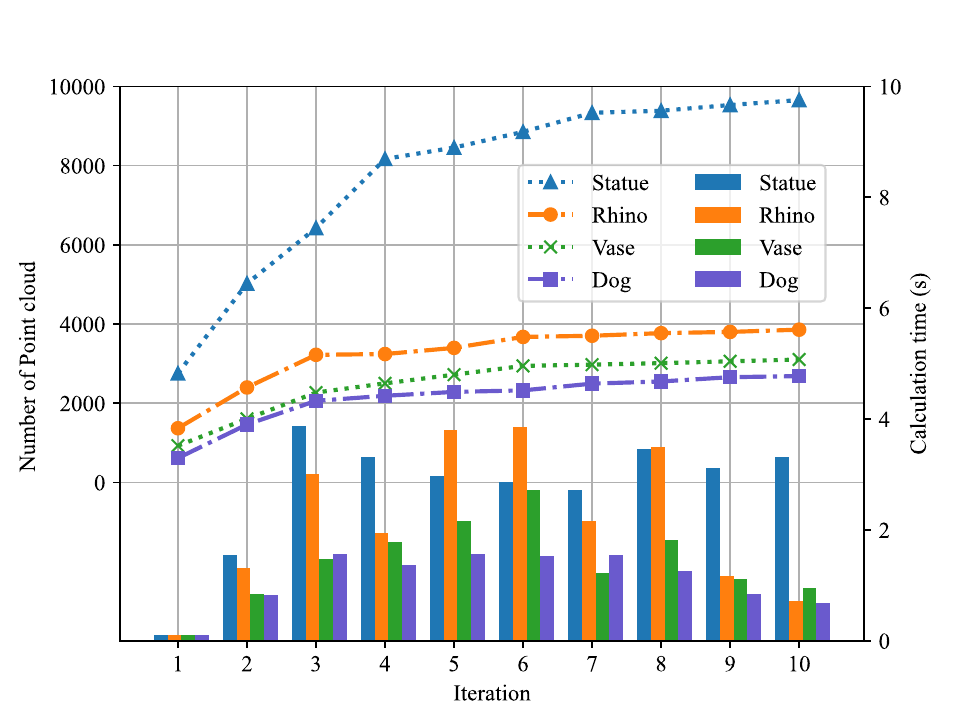}
    \caption{Chart of result data from real-world experiments.The bar chart represents the computational time for each iteration, and the line chart represents the number of points in the reconstructed point cloud after each iteration.}
    \label{Real-world chart}
\end{figure}

The real-world experiment does not have ground truth for objects. 
In order to assess the performance of the framework, we replaced the point cloud coverage metric from the simulation experiment with the number of points in the registered point cloud. 
The number of points in the registered point cloud after filtering with voxels of 0.005m edge length is used to evaluate the convergence efficiency of the framework. 
The overall results are shown in Fig. \ref{Real-world chart}.

Fig. \ref{Real-world chart} illustrates that objects with simple geometric structures (such as the vase and dog in Fig. \ref{Real-world result}) have lower computational times per iteration. 
In contrast, objects with complex structures (such as the Statue and Rhino in Fig. \ref{Real-world result}) require more ellipsoids to describe their geometry, leading to increased computational times. 
Despite the Statue having twice the voxel count of the Rhino, its computational time is similar, indicating that the framework's computational time is not significantly correlated with the size of the object's volume.
\begin{table}[h]
    \caption{Model growth rate}
    \centering
    \begin{tabular}{lllllllll}
    \hline
    Iteration times  & 2 & 3 & 4 & 5 & 6 & 7 & 8 ... \\ 
    \hline
    Statue &0.82&0.28&0.27&\textbf{0.04}&0.05&0.05&0.01 \\ 
    Vase    &0.74&0.41&0.10&\textbf{0.09}&0.08&0.01&0.01\\ 
    Rhino &0.75&0.34&\textbf{0.01}&0.05&0.08&0.01&0.02\\
    dog &1.38&0.41&\textbf{0.06}&0.04&0.02&0.07&0.02&\\
    \hline
    \end{tabular}
    \label{Model growth rate}
\end{table}
Table \ref{Model growth rate} illustrates the increase in point cloud data during the current iteration compared to the previous one. 
The data reveals that since the fifth iteration, the growth rate for all targets has been below 0.1, indicating that the framework possesses high convergence efficiency for targets of diverse sizes and geometric complexities.

\section{Conclusion}


This paper presents a Projection-Based NBV (Next Best View) Planning Framework. The framework has been deployed on a robotic arm equipped with a 3D camera, and experimental results demonstrate that the system can efficiently accomplish the task of complete 3D reconstruction of unknown objects.
The innovation of this paper is that, based on the voxel structure, different types of voxels are fitted into ellipsoids, and the ellipsoids are used to represent the unknown structure. 
The best viewpoint is selected by integrating the weighted projection of ellipsoids of various types with a global partitioning strategy.
The use of this framework greatly reduces the computational burden brought by ray-casting while ensuring the same scanning accuracy. 
This paper also verifies the efficiency and feasibility of this framework through sufficient simulation and real-world experiments. 
Future work will focus on the following aspects: first, using a mobile robotic arm equipped with a 3D camera to complete the 3D reconstruction of large targets, and second, optimizing the motion trajectory of the robotic arm to make the reconstruction process more efficient.

\bibliographystyle{IEEEtran}

\vfill

\end{document}